\documentclass[letterpaper]{article} 
\usepackage{aaai2026}  
\usepackage{times}  
\usepackage{helvet}  
\usepackage{courier}  
\usepackage[hyphens]{url}  
\usepackage{graphicx} 
\usepackage{natbib}  
\usepackage{caption} 
\usepackage{algorithm}
\usepackage{algorithmic}

\usepackage{newfloat}
\usepackage{listings}
\usepackage{amsfonts}       
\usepackage{nicefrac}       
\usepackage{microtype}      
\usepackage{xcolor}         
\usepackage{graphicx}
\usepackage{multirow}
\usepackage{natbib}
\usepackage{subfig}
\usepackage{listings}
\usepackage{adjustbox}
\usepackage[most]{tcolorbox} 
\usepackage{amsmath}
\usepackage{amsthm}  
\usepackage{booktabs}
\usepackage{multicol}

\DeclareCaptionStyle{ruled}{labelfont=normalfont,labelsep=colon,strut=off} 
\floatstyle{ruled}
\newfloat{listing}{tb}{lst}{}
\floatname{listing}{Listing}
\newcommand{\posdiff}[1]{{\textsuperscript{$\uparrow$#1}}}
\newcommand{\negdiff}[1]{{\textsuperscript{$\downarrow$#1}}}
\newcommand{\posdiffinline}[1]{{$\uparrow$#1}}
\newcommand{\negdiffinline}[1]{{$\downarrow$#1}}

\title{Topic Over Source: The Key to Effective Data Mixing for Language Models Pre-training}
\author{
      Jiahui Peng$^{1}$\thanks{Equal contribution.}, 
      Xinlin Zhuang$^{1,2*}$, 
      Jiantao Qiu$^{1*}$, 
      Ren Ma$^{1}$\thanks{Project lead.}, 
      Jing Yu$^{1}$, 
      He Zhu$^{1,3}$, 
      Conghui He$^{1}$\thanks{Corresponding authors.}\\
}
\affiliations{
    \textsuperscript{\rm 1}Shanghai AI Laboratory, \\
    \textsuperscript{\rm 2}School of Computer Science and Technology, East China Normal University, \\
    \textsuperscript{\rm 3}Peking University,\\
    \texttt{{pengjiahui,qiujiantao,maren,heconghui}@pjlab.org.cn},\\
    \texttt{xinlinzhuang@stu.ecnu.edu.cn}
}

\begin{document}
\nocopyright
\maketitle

\begin{abstract}
The performance of large language models (LLMs) is significantly affected by the quality and composition of their pre-training data, which is inherently diverse, spanning various languages, sources, and topics. Effectively integrating these heterogeneous data groups is crucial for optimizing LLM performance. 
Previous research has predominantly concentrated on source-based data mixing, often neglecting the nuanced topic-level characteristics of the data. To address this gap, we propose a \textbf{topic}-based data mixing strategy that utilizes detailed topic labels generated through a multi-stage  process combining unsupervised clustering, LLM-based summarization, and supervised classifier training. 
With this strategy, we conduct the first comprehensive comparison of topic-based versus source-based partitioning across multiple mixing strategies. 
We demonstrate that language models pretrained on data mixed by topics consistently outperform those trained on data mixed by sources across multiple methods including RegMix, DoReMi, temperature-based sampling, and a manual mixing method based on downstream task performance. 
Our theoretical analysis reveals that topic-based data achieves significantly lower validation loss compared to source-based approaches, creating a better optimization landscape for model training. 
We will make our code, annotated datasets, and topic classification models publicly available to facilitate further research.
\end{abstract}

\section{Introduction}
The performance of large language models (LLMs) is profoundly shaped by the quality and composition of their pre-training data \cite{pretrainersguide, dataeverywhere, meta-rater, datasurvey}. 
Existing data mixing strategies range from basic methods such as temperature-based sampling \cite{dataeverywhere} to more advanced approaches including RegMix \cite{regmix}, DoReMi \cite{doremi}, and DoGE \cite{doge}. 
Despite their effectiveness, these techniques primarily function at the source level, viewing each data source (e.g., Wikipedia, GitHub, CommonCrawl) as a uniform collection. 
This source-centric approach fails to recognize an important reality: a single source (e.g., CommonCrawl) may contain multiple topics of different relevance to specific downstream tasks, and likely the same topic (e.g., Science) can appear across multiple sources with varying quality and presentation styles. 
Moreover, the growing trend towards web-crawled datasets further diminishes the utility of source-based mixing. 
Recent efforts like FineWeb \cite{fineweb} and DCLM \cite{DCLM2024} produce multi-trillion-token datasets predominantly from CommonCrawl, offering no meaningful source divisions. 
In this context, semantic organization becomes not just beneficial but necessary for effective data curation.
\par
An alternative data curation paradigm, which organizes data based on intrinsic semantic properties instead of provenance, has therefore gained prominence. Pioneering methods have demonstrated the potential of this approach, but they have been constrained by methodological limitations. For example, WebOrganizer \cite{WebOrganizer} proposed structured taxonomies that required substantial human supervision, limiting both scalability and generalization. In contrast, frameworks like R\&B \cite{ge2025r} employed unsupervised clustering for automatic domain discovery, but this process produces unlabeled groupings and its effectiveness has not been demonstrated on web-scale pre-training corpora. While these foundational contributions validate the principle of semantic partitioning, they reveal a critical gap: there is currently no fully-automated and scalable methodology for producing a coherent, high-quality topic taxonomy from web-scale data.
\par
In response to these limitations, we present a comprehensive empirical evaluation of partitioning strategies based on semantic content through a systematic framework.
 First, our approach implements a multi-stage clustering methodology to identify detailed topics within extensive datasets. 
We cluster semantically related documents using embedding techniques, followed by applying LLMs to create descriptive topic labels that effectively capture each cluster's semantic essence. 
This method produces a comprehensive taxonomy offering deeper insights into dataset structure compared to conventional source-based classifications, providing a more refined strategy for pre-training data composition that naturally aligns with the semantic foundations of content that LLMs process.
\par
Second, using this framework, we conduct the first large-scale, comprehensive comparison of topic-based versus source-based partitioning across multiple established mixing algorithms, including performance-based reweighting (PerfRe), temperature-based sampling, DoReMi, and RegMix. Our empirical results show that our topic-based mixing strategy consistently yields better results than source-based approaches. These performance gains remain consistent when tested with larger models and extended training sequences, validating the effectiveness of our approach.
The main contributions of this paper are as follows:

\begin{itemize}
    \item We propose a scalable topic extraction method at web-scale that combines unsupervised clustering, LLM-based summarization, and supervised classifier training to partition the SlimPajama dataset into 12 semantically meaningful topics. This annotated dataset will be open-sourced to facilitate further research. 
    \item We conduct the first comprehensive comparison of topic-based versus source-based data partitioning across multiple established mixing algorithms and different model scales. Our results provide robust evidence for the superiority of the topic-based approach.
    \item We show that topic-based organization of training data improves the optimization landscape, creating a better relationship between mixture weights and model performance than source-based approaches, leading to more effective training configurations. 
\end{itemize}

\section{Topic Extraction}
\label{sec:Topic Extraction}

\begin{algorithm}[t]
    \caption{Topic Extraction Process}
    \label{alg:topic_extract}
    \begin{algorithmic}[1]
    \REQUIRE Dataset $\mathcal{D}$ with $N$ documents, parameters $k_1$, $k_2$, $m_{topic}$
    \STATE \textbf{Output:} Topic taxonomy with $m_{topic}$ topics and topic labels for all documents
    
    \STATE // Step 1. Unsupervised Clustering
    \STATE Generate embeddings $E = \{e_1, e_2, \dots, e_N\}$ for all documents using BGE model
    \STATE Apply K-Means to partition $E$ into $k_1$ clusters: $C_1, C_2, \dots, C_{k_1}$
    \FOR{$i = 1$ to $k_1$}
        \STATE Sample 10 documents from cluster $C_i$
        \STATE Generate summary $S_i$ using gpt-4o for the sampled documents
    \ENDFOR
    \STATE Apply K-Means to group the $k_1$ centroids into $k_2$ clusters
    \FOR{$j = 1$ to $k_2$}
        \STATE Sample 50 summaries from $S_i$ for cluster $j$
        \STATE Generate abstract topic $T_j$ using gpt-4o
    \ENDFOR
    
    \STATE // Step 2. Topic Extraction with LLM
    \STATE Merge $k_2$ topics into $m_{topic}$ final topics using gpt-4o
    \STATE Obtain final topic taxonomy $\mathcal{T} = \{T_1, T_2, \dots, T_{m_{topic}}\}$
    
    \STATE // Step 3. Training Topic Classifier
    \STATE Sample 100,000 documents from $\mathcal{D}$
    \STATE Annotate topics for sampled documents using gpt-4o and topic taxonomy $\mathcal{T}$
    \STATE Train a BERT-based classifier $\mathcal{M}$ on the annotated data
    \STATE Use $\mathcal{M}$ to classify all documents in $\mathcal{D}$ into topics in $\mathcal{T}$
    
    \STATE \textbf{Return:} Topic taxonomy $\mathcal{T}$ and topic labels for all documents in $\mathcal{D}$
    \end{algorithmic}
\end{algorithm}

\subsection{Dataset}
We employ SlimPajama \cite{slimpajama}, a widely recognized dataset containing 600 million documents with roughly 627B Llama tokens. The corpus is organized into seven source categories: arXiv, Books, C4, CommonCrawl, GitHub, StackExchange, and Wikipedia.

\subsection{Topic Extraction Procedure}
Given the large scale of the SlimPajama dataset, we design a multi-stage method to extract the topics. 
The workflow of our topic extraction process is provided in Algorithm \ref{alg:topic_extract}.

\paragraph{Unsupervised Clustering.}
The initial phase involves generating semantic embeddings for all 600 million documents using the BGE model \cite{bge_embedding}. 
We then apply K-Means clustering to partition these embeddings into $k_1$ distinct clusters. 
From each cluster, we randomly select 10 examples and utilize gpt-4o\footnote{We use \textit{gpt-4o-2024-11-20} throughout the paper.} to create a concise summary capturing the cluster's essential characteristics. The sample size of 10 was chosen to balance comprehensive representation with the constraints of gpt-4o's context window and the typical length of documents and summaries.

In the subsequent phase, we perform a second-level clustering by applying K-Means to group the $k_1$ clusters into $k_2$ higher-level clusters. 
For each of these $k_2$ clusters, we randomly sample 50 of the previously generated summaries and prompt gpt-4o to synthesize an abstract topic label, ultimately yielding $k_2$ distinct topics. 
We implement this two-stage K-Means approach to accommodate both the scale of our dataset and our computational limitations. Our approach aligns with previous work on scalable clustering for large datasets \cite{johnson2019billion, meng2015mllibmachinelearningapache}.

\paragraph{Topic Extraction with LLM.}
Despite the effectiveness of combining K-Means clustering with gpt-4o for identifying key topics, our analysis reveals limitations in the unsupervised approach. The clustering process generated considerable topic overlap, with our manual evaluation of the $k_2$ topics uncovering redundancies where multiple topics shared similar semantic elements. We also identify inconsistencies in topic granularity—some topics are narrowly defined while others remain overly broad—which compromises the interpretability of our taxonomy. These challenges, documented with examples in Appendix \ref{app:merging_topics}, necessitated an additional consolidation step. To address this, we leveraged gpt-4o to merge the $k_2$ topics into $m_\text{topic}$ final topics, resulting in a more refined and coherent topic structure. Following extensive experimentation, we established optimal cluster parameters of $k_1=10,000$, $k_2=300$, and $m_\text{topic}=12$ for our implementation. The detailed prompts for both summary generation and topic consolidation are available in Appendix \ref{app:prompt}.

\paragraph{Topic Classifier Training.}
Following the previous steps, we establish a comprehensive topic taxonomy that divides the dataset into 12 distinct topics. However, manual inspection of randomly selected samples reveals that document topics assigned through unsupervised clustering sometimes misalign with their actual content. Additionally, the clustering approach lacks scalability to new documents, whereas a classifier can efficiently process additional data with similar distributions to SlimPajama.
To address these limitations, we leverage our derived taxonomy to create a supervised classifier. 
We randomly sample 100,000 documents from SlimPajama and use gpt-4o to annotate their topics according to our taxonomy. 
Using these annotations, we fine-tune a BERT classifier to effectively distill gpt-4o's topic classification capabilities. 
The classifier achieves 84\% accuracy on our test set and is subsequently deployed to categorize the entire SlimPajama dataset. 
Details about classifier architecture, and fine-tuning procedures are available in Appendix \ref{app:topic_classifier}.

\begin{figure}[tb]
    \centering
    \includegraphics[width=0.8\linewidth]{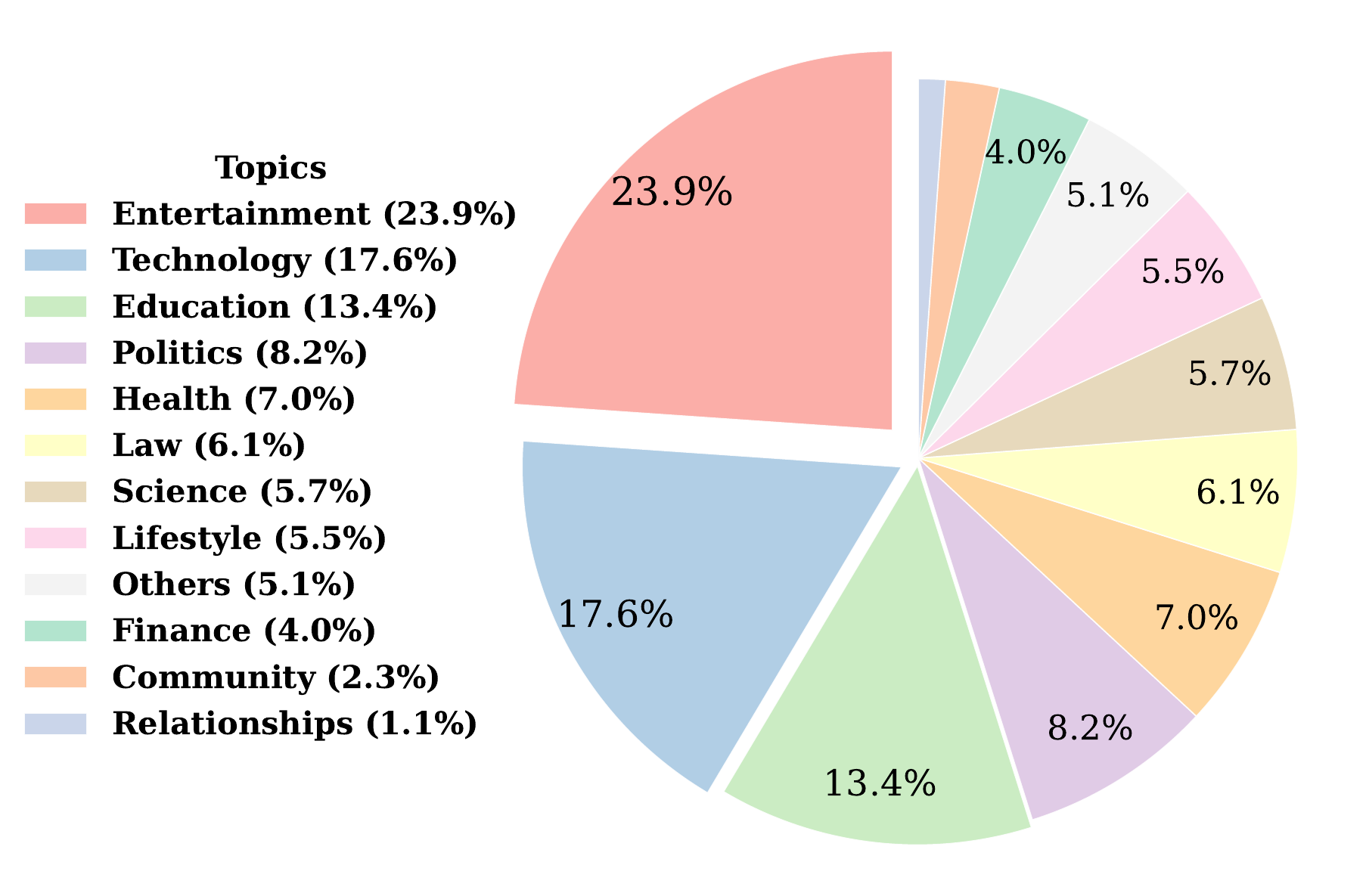}
    \caption{Topic analysis of the SlimPajama dataset. Distribution of 12 topics showing significant imbalance, with Entertainment (23.9\%) and Technology (17.6\%) comprising over 41\% of the content.}
    \label{fig:topic_distribution_a}
\end{figure}

\begin{figure}[tb]
    \centering
    \includegraphics[width=0.8\linewidth]{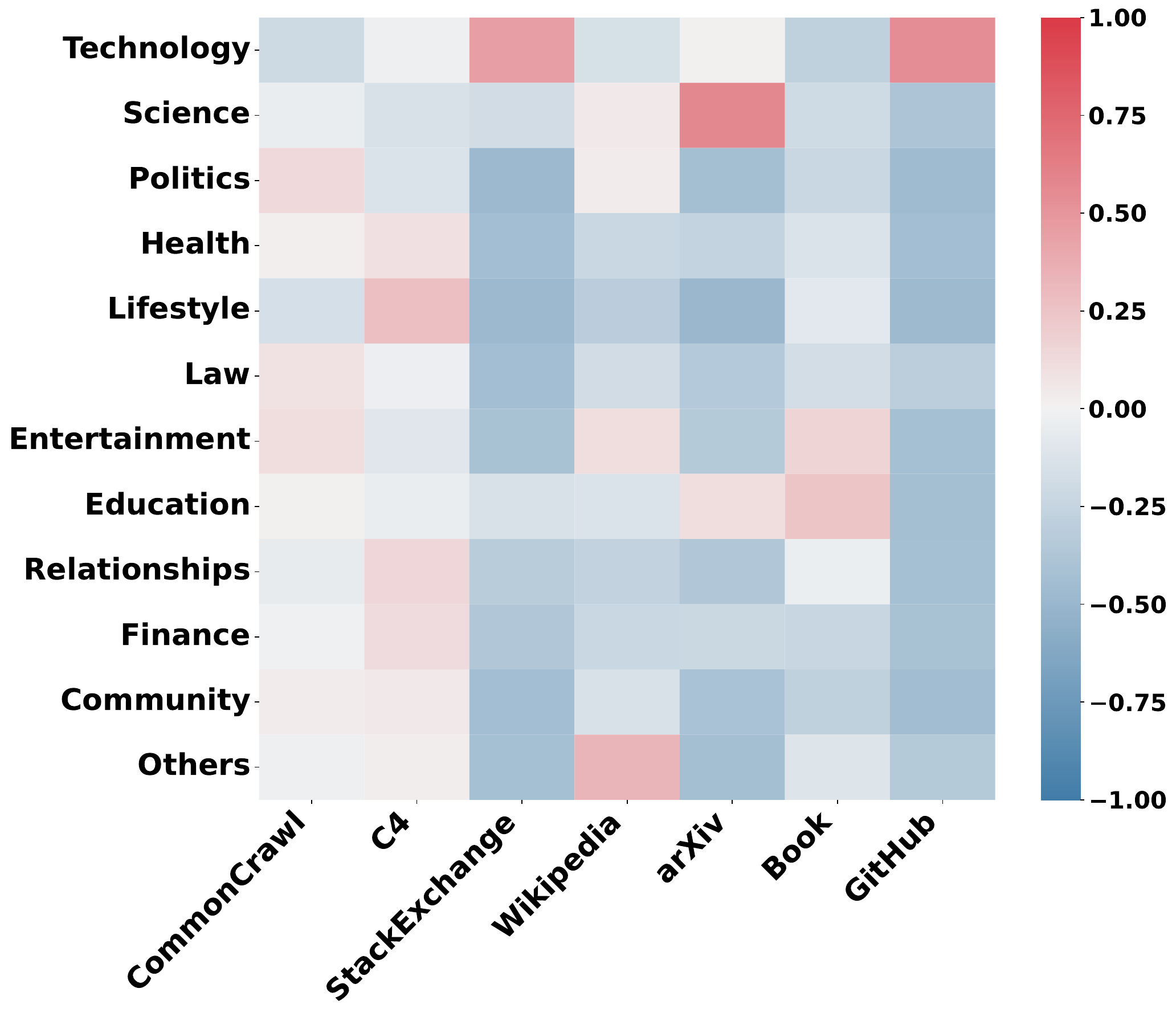}
    \caption{NPMI heatmap between topics and sources, where red indicates strong association, blue shows mutual exclusivity, and white represents minimal association, highlighting the complementary information provided by these two dimensions.}
    \label{fig:topic_distribution_domain}
\end{figure}

\subsection{Topic Distribution}
Our analysis of SlimPajama extracts 12 distinct topics: Technology, Science, Politics, Health, Lifestyle, Law, Entertainment, Education, Relationships, Finance, Community, and Others. These categories align well with both the Wikipedia ontology\footnote{\url{https://en.wikipedia.org/wiki/Wikipedia:Contents/Categories}} and the manually crafted taxonomy in \cite{WebOrganizer}, demonstrating our method's effectiveness in identifying meaningful and human-interpretable topics.
\par
Figure \ref{fig:topic_distribution_a} shows significant topic imbalance in the dataset. Entertainment (23.9\%) and Technology (17.6\%) dominate, together making up over 41\% of content. Education follows at 13.4\%, with Politics (8.2\%), Health (7.0\%), and Law (6.1\%) moderately represented. Science comprises only 5.7\% despite its importance, while social domains like Relationships (1.1\%) and Community (2.3\%) are least represented. 
This imbalance likely causes models to perform better on technology-related tasks while struggling with scientific reasoning and social contexts.
Additional examples for our topic categorization are provided in Appendix \ref{app:topic_extraction_case_study}.

\subsection{Relationship Between Topic and Source}
To explore the relationship between topic and source in the dataset, we calculate the normalized pointwise mutual information (NPMI) between these two dimensions, where values near zero suggest independence, positive values indicate association, and negative values reflect mutual exclusivity.
Our examination of the NPMI matrix across SlimPajama data sources, illustrated in Figure \ref{fig:topic_distribution_domain} right, reveals a varied landscape of relationships between topics and sources.
While some natural associations exist—such as ArXiv's positive relationship with Science—many topic-source combinations display values suggesting these dimensions capture different aspects of the content. 
For instance, topics like Law, Health, and Education show different patterns of association across sources like CommonCrawl and Wikipedia.
This diversity in relationships highlights how topic and source dimensions provide complementary information about the data. This complementary nature is particularly valuable for downstream applications, as it indicates that considering both dimensions would yield more nuanced data characterization, enabling more diverse and balanced dataset construction strategies.

\section{Data Mixing for Pre-training}

\subsection{Task Formulation}
In our framework, we work with a dataset $\mathcal{D} = \{\mathcal{D}_1, \ldots, \mathcal{D}_m\}$ divided into $m$ distinct groups. 
Data mixing involves finding an optimal weight vector $p = \left[p_1, \ldots, p_m\right] \in \triangle^m$ on the probability simplex. The primary objective is to optimize a language model $\pi_\theta | p$ for downstream performance by minimizing validation loss: 
\begin{equation}
    \underset{p \in \triangle^{m}}{\text{minimize}} \sum_{i = 1}^m \mathcal{L}_{\text{val},i}(\pi_\theta | p).
\end{equation}
where $\mathcal{L}_{\text{val},i}(\pi_\theta | p)$ denotes the validation loss of the model $\pi_\theta | p$ on the $i_{th}$ data group after pre-training.
While conventional data mixing strategies typically group datasets by their sources with $m_\text{source}$ total sources, our approach introduces a novel dimension by grouping data according to semantic content. 
We organize data into topics (with $m_\text{topic}$ total topics), representing meaningful categories such as Science and Lifestyle that capture the semantic essence of the content.

\subsection{Experimental Setup}
\label{subsec:setup}

\paragraph{Training.}
The models utilized in this work are 1.3B parameter decoder-only transformers. 
Key architectural features include Rotary Position Embeddings (RoPE) \cite{su2024roformer} and a maximum context window of 1,024 Llama tokens \cite{llama}. 
For the standard pre-training configuration, each model is trained from scratch on a selected subset of 30B tokens. 
A comprehensive description of the architecture and training setup is available in Appendix \ref{app:pretrain_detail}.

\paragraph{Evaluation.}
To evaluate the capabilities of pre-trained LLMs, we assess their performance through in-context learning using the \textit{lm-evaluation-harness} framework \cite{eval-harness}, with accuracy scores reported as the performance metric. 
The evaluation dataset spans three distinct categories of downstream tasks: (1) \textbf{General Knowledge}: ARC-Challenge \cite{arc}, ARC-Easy, and SciQ \cite{sciq}; (2) \textbf{Commonsense Reasoning}: PIQA \cite{bisk2020piqa}, SIQA \cite{siqa}, WinoGrande \cite{winogrande}, and CommonsenseQA \cite{commonsenseqa}; (3) \textbf{Reading Comprehension}: RACE \cite{race} and OpenBookQA \cite{openbookqa}.
Further details regarding the evaluation process are provided in Appendix \ref{app:evaluation}.

\subsection{Baselines}
In this paper, we pre-train models with the following data mixing methods. 
The specific data weights employed across all experimental settings can be found in Appendix \ref{app:all_mixing_weights}.

\paragraph{Random.} This serves as a general baseline method. Tokens are randomly sampled from the training set of the raw SlimPajama dataset, without applying any specific control over the data distribution. In this case, the data group proportions strictly adhere to the token distribution of data groups. 
\paragraph{PerfRe.} In the practice of pre-training LLMs, researcher often manually adjust sampling ratios based on downstream performance \cite{dubey2024llama, gopher}. 
To benchmark against such approaches, we introduce \textbf{Perf}ormance-based \textbf{Re}weighting (\textbf{PerfRe}), a systematic manual data mixing strategy. 
This method involves systematically upsampling individual data groups (either topics or sources) and evaluating their impact on downstream task performance. 
We then prioritize data groups that yield the greatest performance improvements. 
Following methodology similar to Llama-3.1 \cite{dubey2024llama}, we start with the Random baseline model trained on 30B tokens, then upsample each selected data group by 30\% (while normalizing the remaining groups' ratios) and conduct continual pre-training for additional 30B tokens with this modified mixture. 
This process generates $m_\text{topic}$ models (one per upsampled topic) and $m_\text{source}$ models (one per upsampled source). 
Based on performance analysis, we identify most beneficial topics (Science, Relationships, and Health) or sources (CommonCrawl and C4), upsample them by 30\% (normalizing the remaining groups), and pre-train new models from scratch for 30B tokens using these optimized mixtures. 
Fulll evaluation results of models trained using PerfRe weights are available in Appendix \ref{app:continual_pretrain}.
\paragraph{Temperature.} Temperature-based sampling \cite{dataeverywhere,bert} proportionally adjusts data source weights according to a scaled factor of their token counts. 
We set $t=0.4$ to compute topic weights based on token ratios.
\paragraph{RegMix.} RegMix \cite{regmix} involves training a set of small 1M-parameter models on diverse data mixtures and fitting regression models using lightGBM to predict model performance based on the respective mixtures. Using the fitted regression model, the top-ranked mixture is simulated to determine the optimal topic weights.
\paragraph{DoReMi.} DoReMi \cite{doremi} employs a small proxy model trained using group distributionally robust optimization (Group DRO) to generate domain weights.
\par
For a clear comparison between topic-based and source-based data mixing strategies, we separately implement and train distinct models using PerfRe, Temperature, RegMix and DoReMi methods with both approaches.

\subsection{Pre-training Results}
\begin{table*}[!tb]
\centering
\small
\begin{tabular}{@{}lcccc@{}}
\toprule
\textbf{Data Mixing Method} &
  \begin{tabular}[c]{@{}c@{}}\textbf{General Knowledge}\\ (3 tasks)\end{tabular} &
  \begin{tabular}[c]{@{}c@{}}\textbf{Commonsense Reasoning}\\ (4 tasks)\end{tabular} &
  \begin{tabular}[c]{@{}c@{}}\textbf{Reading Comprehension}\\ (2 tasks)\end{tabular} &
  \begin{tabular}[c]{@{}c@{}}\textbf{Average}\\ (9 tasks)\end{tabular} \\ \midrule
Random             & 54.52 & 44.42 & 25.07 & 43.49 \phantom{\posdiff{0.00}} \\ \midrule
PerfRe-Source      & 55.27 & 45.85 & 26.26 & 44.63 \posdiff{1.14}           \\
PerfRe-Topic       & 56.36 & 46.23 & 26.52 & 45.23 \posdiff{1.74}           \\ \midrule
Temperature-Source & 53.64 & 45.47 & 25.76 & 43.81 \posdiff{0.32}           \\
Temperature-Topic  & 55.62 & 44.96 & 27.66 & 44.67 \posdiff{1.18}           \\ \midrule
RegMix-Source      & 53.77 & 45.74 & 25.38 & 43.89 \posdiff{0.40}           \\
RegMix-Topic       & 54.39 & 45.96 & 26.16 & 44.39 \posdiff{0.90}           \\ 
\midrule
DoReMi-Source      & 54.36 & 45.34 & 27.19 & 44.31 \posdiff{0.82}           \\
DoReMi-Topic       & 54.98 & 45.50 & 29.02 &  45.00\posdiff{1.51}           \\ \midrule
3.3B Random        & 61.22 & 47.33 & 34.19 & 49.04 \phantom{\posdiff{0.00}} \\
3.3B RegMix-Source & 61.29 & 47.59 & 34.69 & 49.29 \posdiff{0.25}           \\
3.3B RegMix-Topic  & 61.78 & 48.70 & 35.20 & 50.06 \posdiff{1.02}           \\ \bottomrule
\end{tabular}
\caption{Performance of pre-trained models with different data mixing methods on downstream tasks.  The \posdiffinline{x.xx} and \negdiffinline{x.xx} values indicate positive and negative differences compared to the Random baseline, respectively. For 3.3B models, differences are relative to the 3.3B Random baseline. Full results are provided in Appendix \ref{app:full_pretrain}.}
\label{tab:topic_over_source}
\end{table*}

\paragraph{Topic-based data mixing outperforms source-based data mixing.}
As shown in Table \ref{tab:topic_over_source}, the results demonstrate that topic-based data mixing consistently outperforms source-based data mixing across all four methods: PerfRe, Temperature, RegMix and DoReMi. For PerfRe, the topic-based approach achieves an average score of 45.23 compared to 44.63 for source-based mixing, representing a 0.60 point improvement. Similarly, Temperature-Topic outperforms Temperature-Source by 0.86 points (44.67 vs. 43.81), while RegMix-Topic surpasses RegMix-Source by 0.50 points (44.39 vs. 43.89). DoReMi-Topic outpaces DoReMi-Source by 0.69 points (45.00 vs. 44.31).
This consistent pattern of improvement across different mixing strategies suggests that organizing data by semantic content rather than source provides more meaningful signals for model training. Looking at specific task categories, we observe that topic-based mixing particularly excels in Reading Comprehension tasks, where Temperature-Topic achieves a score of 27.66 compared to 25.76 for Temperature-Source, representing a substantial 1.90 point improvement, while DoReMi-Topic similarly outperforms DoReMi-Source with 29.02 versus 27.19, a notable 1.83 point gain. In General Knowledge tasks, PerfRe-Topic shows the largest gain over its source-based counterpart (56.36 vs. 55.27, a 1.09 point difference). These results indicate that semantic organization of training data enables the model to develop stronger representations of knowledge domains and reasoning capabilities.

\paragraph{Scaling to larger models and datasets.}
To verify the effectiveness of topic-based data mixing at larger scales, we conducted experiments with 3.3B parameter models trained on 70B tokens. Notably, the improvement gap between topic-based and source-based mixing methods widens as we scale up, increasing from 0.5 in the 1.3B setting to 0.7 in the 3.3B setting. This increasing advantage demonstrates the enhanced effectiveness of semantic organization at larger scales. As shown in Table \ref{tab:topic_over_source}, the 3.3B RegMix-Topic model achieves an average score of 50.06 across all tasks, substantially outperforming both the 3.3B RegMix-Source model (49.29) and the 3.3B Random baseline (49.04). These improvements confirm that organizing training data by semantic content rather than source is a robust approach that becomes increasingly beneficial as model size and training data grow.

\paragraph{PerfRe shows superior performance among data mixing methods.}
Table \ref{tab:topic_over_source} shows PerfRe outperforming all other data mixing methods.
PerfRe-Topic's average score of 45.23 exceeds DoReMi-Topic (45.00), RegMix-Topic (44.39), and Temperature-Topic (44.67),
with notable advantages in General Knowledge (56.36) and Commonsense Reasoning (46.23) tasks. PerfRe succeeds by systematically identifying and prioritizing beneficial data groups, upsampling valuable topics like Science and Health while balancing others. This approach, similar to findings in Llama-3.1 \cite{dubey2024llama}, demonstrates that strategic data proportion adjustments enhance model capabilities. Our results suggest that well-designed heuristic approaches can be as effective as more complex optimization methods for data mixing.

\section{Analysis}

\begin{table*}[!tb]
\centering
\small
\begin{tabular}{@{}lcccc@{}}
\toprule
\textbf{Data Mixing Method} &
  \begin{tabular}[c]{@{}c@{}}\textbf{General Knowledge}\\ (3 tasks)\end{tabular} &
  \begin{tabular}[c]{@{}c@{}}\textbf{Commonsense Reasoning}\\ (4 tasks)\end{tabular} &
  \begin{tabular}[c]{@{}c@{}}\textbf{Reading Comprehension}\\ (2 tasks)\end{tabular} &
  \begin{tabular}[c]{@{}c@{}}\textbf{Average}\\ (9 tasks)\end{tabular} \\ \midrule
Random                & 54.52 & 44.42 & 25.07 & 43.49 \phantom{\posdiff{0.00}} \\ \midrule
RegMix-Source         & 53.77 & 45.74 & 25.38 & 43.89 \posdiff{0.40}           \\
RegMix-Topic          & 54.39 & 45.96 & 26.16 & 44.39 \posdiff{0.90}           \\ \midrule
RegMix-Topic * Source & 56.56 & 45.05 & 25.65 & 44.58 \posdiff{1.09}           \\ \midrule
FineWeb-Edu           & 56.06 & 45.53 & 25.22 & 44.53 \posdiff{1.04}           \\
\quad + RegMix-Source & 57.91 & 46.02 & 26.77 & 45.70 \posdiff{2.21}           \\
\quad + RegMix-Topic  & 58.10 & 46.01 & 27.88 & 46.01 \posdiff{2.52}           \\ \midrule
RegMix-Topic-7        & 56.06 & 45.53 & 25.22 & 44.53 \posdiff{1.04}           \\ \bottomrule
\end{tabular}
        \caption{Performance of ablation analysis of data mixing strategies including (1) combining topic and source dimensions, (2) integrating data mixing with FineWeb-Edu, and (3) varying topic counts.
     The top three rows repeat results from Table \ref{tab:topic_over_source} for easier comparison. Full results are provided in Appendix \ref{app:full_ablation}.}
    \label{tab:ablation_results}
\end{table*}

\subsection{Integrating Topic and Source Dimensions in Data Mixing}
Figure \ref{fig:topic_distribution_domain} reveals varied relationships between topics and sources, with each dimension providing complementary information about the data. 
This raises the question of whether combining topic and source information could enhance data mixing strategies. 
Inspired by WebOrganizer \cite{WebOrganizer}, we independently calculate the proportions for each topic and source using RegMix, then merge them through a Cartesian product operation. 
This approach created $m_\text{topic} \times m_\text{source}$ distinct groups. 
We then pre-trained a model from scratch using 30B tokens sampled from these combined groups and evaluated its performance on downstream tasks. As shown in Table \ref{tab:ablation_results}, this integrated approach (RegMix-Topic * Source) outperforms both pure RegMix-Topic (44.39) and RegMix-Source (43.89), achieving an average score of 44.58 across all tasks. 
This represents a 1.09 point improvement over the random baseline, demonstrating the value of leveraging both dimensions for data mixing.

\subsection{Combining Data Mixing with Data Selection}
Data selection involves filtering a large dataset to extract a subset meeting specific criteria \cite{meta-rater}.
While this technique is commonly used independently, its combination with data mixing during pre-training remains relatively unexplored.
We investigate this combined approach in our research.
Our methodology involves first determining token allocations for each group using RegMix, then employing the established FineWeb-Edu classifier \cite{fineweb} to select the appropriate number of tokens per group, followed by pre-training a model from scratch with this curated dataset. We implement this strategy for both topic-based and source-based data mixing approaches.
The results in Table \ref{tab:ablation_results} demonstrate that this integrated method yields an average score of 46.01 for topic-based data mixing (FineWeb-Edu + RegMix-Topic), outperforming both the FineWeb-Edu + RegMix-Source combination (45.70) and the standalone FineWeb-Edu classifier method (44.53). These findings indicate that combining data selection with data mixing techniques can enhance pre-trained model performance, with topic-based mixing providing the greatest benefit (2.52 points improvement over the random baseline).

\subsection{Effect of the Number of Topics}
In the experiments comparing topic- and source-based data mixing in Table \ref{tab:topic_over_source}, we use different numbers of groups: $m_\text{topic}=12$ for topic-based mixing vs. $m_\text{source}=7$ for source-based mixing. 
To ensure a fair comparison and investigate whether the performance difference stems from the grouping method itself rather than the number of groups, we further merge our 12 topics into 7 using gpt-4o (despite creating a suboptimal topic taxonomy), compute the mixing ratios using RegMix, and pre-train a language model (RegMix-Topic-7) on this new data mixture. 
As shown in Table \ref{tab:ablation_results}, the RegMix-Topic-7 achieves an average score of 44.53 across all tasks, which is 1.04 points higher than Random. 
This performance is similar to our original 12-topic model (RegMix-Topic, 44.39), with both topic-based approaches surpassing the source-based method (RegMix-Source, 43.89). 
These results demonstrate that the benefits of topic-based mixing derive primarily from the semantic organization of data rather than simply from having a larger number of groups.

\section{Discussion}
From a theoretical perspective, data mixing can be formulated as an optimization task to find the optimal weight vector $p$ that minimizes the validation loss $\mathcal{L}_{\text{val}}$ of $\pi_\theta | p$. Data mixing methods essentially solve an optimization problem that determines proportions to minimize the validation loss, based on an implicit functional relationship between group losses and mixture proportions. We can denote this mapping from mixture proportions to loss as $f$:

\begin{equation}
\mathcal{L}_{\text{val}} (\pi_\theta | p) = f(p)
\end{equation}
\begin{figure*}[!t]  
    \centering
    \includegraphics[width=0.95\textwidth]{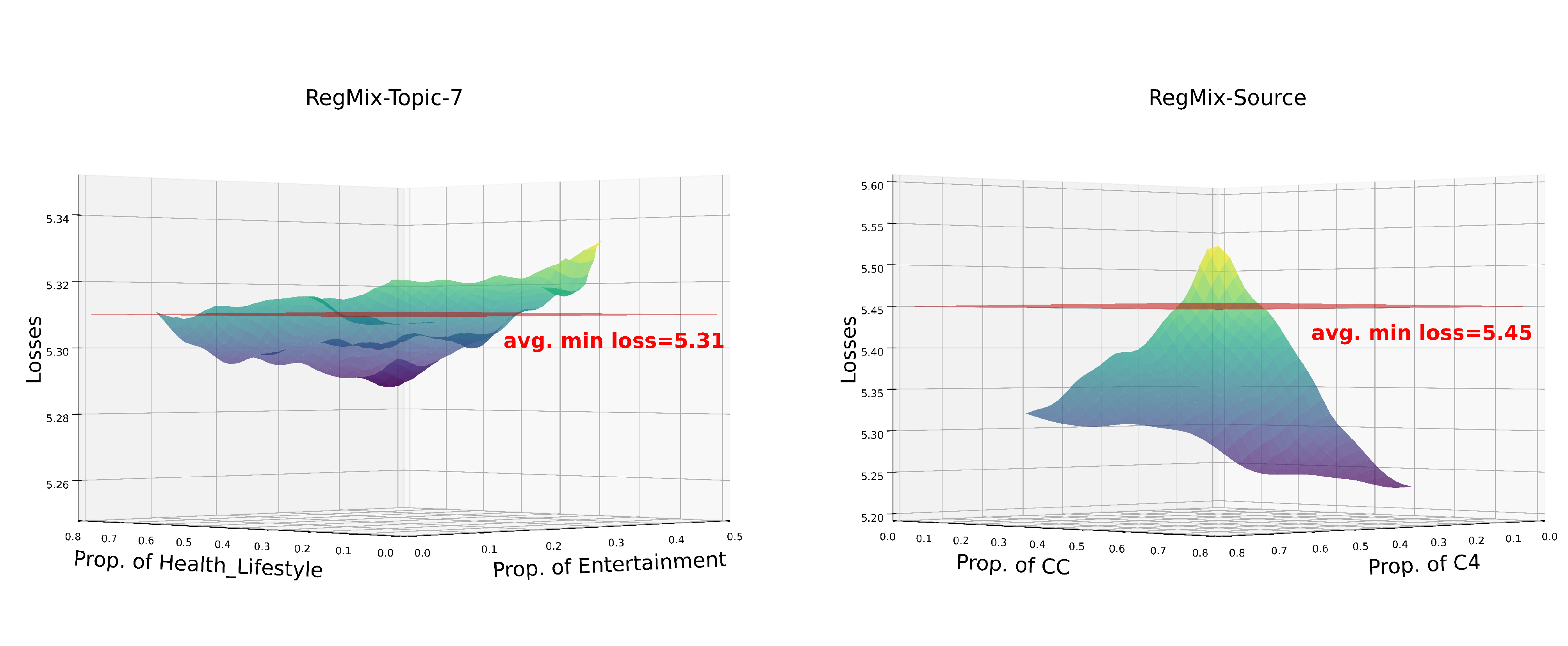} 
    \caption{Two 3D loss landscapes comparing different mixture strategies. \textbf{(Left)} RegMix-Topic-7 shows loss variation across Health\_Lifestyle and Entertainment topic proportions. \textbf{(Right)} RegMix-Source shows loss variation across CommomCrawl (CC) and C4 data source proportions. RegMix-Topic-7 achieves a lower averaged minimum loss (5.31) than RegMix-Source (5.45), demonstrating a 0.14 points improvement when organizing data by semantic content rather than by source.}
    \label{fig:loss_topic_source}
\end{figure*}
Different data mixing approaches provide distinct solutions for finding optimal mixture weights through this implicit mapping: DMLaws employs a fixed exponential formula to predict performance based on data ratios. DoReMi implements a dynamic process that compares performance against a reference model trained with equal proportions, adjusting weights for under-performing data groups and averaging these proportions for the final model. RegMix fits this mapping using a lightGBM model with data collected from multiple proxy models.
\par
For the difference between topic and source in data mixing, we argue that the primary distinction lies in the achievable minimum validation loss under different mapping functions. When optimizing mixture weights, the underlying relationship between proportions and performance differs substantially depending on whether data is organized by topic or by source. Topic-based mixing appears to provide a more favorable optimization landscape, enabling the discovery of weight configurations that yield lower validation losses.

Using RegMix as a concrete example, when a lightGBM model is fitted using proxy models trained with topic-based mixing, it can more effectively model the relationship between mixture weights and model performance. This results in finding better weight combinations that lead to lower overall loss compared to source-based mixing. The topic-based organization likely creates more coherent and semantically meaningful data groupings that allow for more efficient knowledge transfer during training, whereas source-based groupings may contain more heterogeneous content that is harder to optimize jointly.
To validate this interpretation, we evaluate the predictive capabilities of lightGBM models trained in both RegMix-Topic-7 and RegMix-Source configurations. 
By generating 100,000 random weight vectors $\mathbf{\tilde{w}}$ across an expanded parameter space, we predict their corresponding losses using $\hat{l} = f(\mathbf{\tilde{w}})$. We take the average of the lowest half losses for robustness. We find that the averaged minimum loss of RegMix-Topic-7 (5.31) is significantly lower than that for RegMix-Source (5.45), with a (0.14) points gap. To better visualize this difference, we plot the loss distribution along two selected topics/sources in Figure \ref{fig:loss_topic_source}.
This suggests that the semantic organization of training data (by topic) provides a more effective basis for optimization than organizational provenance (by source), leading to superior performance in the final model.

\section{Related Work}

\paragraph{Data Curation.}
The quality of pre-training data is critical for model performance \cite{pretrainersguide, dataeverywhere}. While data mixing strategies like DoReMi \cite{doremi} and RegMix \cite{regmix} are effective, their reliance on pre-defined sources is a growing limitation for web-scale corpora. This has spurred a trend towards constructing semantic domains directly from the data. Pioneering works have successfully demonstrated the potential of this approach. For instance, WebOrganizer \cite{WebOrganizer} introduced well-behaved taxonomies for topic and format, while frameworks like R\&B \cite{ge2025r} automated domain discovery using unsupervised clustering. However, these studies primarily focused on validating the effectiveness of their own specific methodologies and thus did not provide the evidence of semantic partitioning superiority over source-based mixing.

\paragraph{Topic Extraction.}
The landscape of topic extraction methods encompasses both unsupervised and heuristic approaches. 
Unsupervised techniques, such as the Latent Dirichlet Allocation (LDA)\cite{blei2003latent} and BERTopic\cite{grootendorst2020bertopic}, typically generate lists of topic words to represent topics in large text corpora. \cite{rijcken2023towards}. \cite{mu2024addressing, mu2024large, rijcken2023towards}  employed LLMs to extract coherent topics on small dataset limited by annotation costs.
In the pre-training task, High-quality taxonomies have often depended on significant human intervention, such as the manual design and specialized knowledge required in approaches \cite{llama,WebOrganizer}. 
The reliance on human-in-the-loop processes for data curation presents significant challenges in scalability and generalization, rendering such methods impractical for web-scale pre-training corpora. Motivated by this gap, our work introduces a novel and scalable topic extraction pipeline. This method integrates three key stages: unsupervised clustering, LLM-based summarization, and the training of a supervised classifier. Applying this pipeline, we partition the SlimPajama dataset into 12 semantically meaningful topics.

\section{Limitations}



A notable limitation is our experiment scale (1B models, 30B tokens), where data curation impacts may not be fully apparent \cite{meta-rater, QuRating}. Nevertheless, we hypothesize these results represent a lower bound. This is supported by findings that data mixing effects amplify with scale \cite{regmix} and our preliminary 3B model results, which showed greater performance gains. We thus expect the benefits to be more substantial for state-of-the-art models and web-scale data, a clear avenue for future work.

\section{Conclusion}

This study introduces a novel topic-based data mixing strategy for language model pre-training that consistently outperforms traditional source-based approaches across multiple mixing methods, model sizes, and training tokens. By combining unsupervised clustering, LLM-based summarization, and supervised classification, we effectively partition training data into semantically meaningful topics that provide more valuable signals for model training than source provenance.
\par
Our comprehensive experiments provide the first large-scale, systematic evidence of this approach's superiority, and demonstrate that topic-based organization creates a more favorable optimization landscape, yields better downstream task performance, and shows increasing benefits when scaling to larger models. This superiority holds true even when combined with techniques like data selection. These findings provide clear evidence that understanding the semantic structure of pre-training data is fundamental to developing more capable language models, offering practitioners a practical and scalable approach to maximize downstream performance.

\bibliography{aaai2026}

\begin{thebibliography}{35}
\providecommand{\natexlab}[1]{#1}

\bibitem[{Albalak et~al.(2024)Albalak, Elazar, Xie, Longpre, Lambert, Wang, Muennighoff, Hou, Pan, Jeong, Raffel, Chang, Hashimoto, and Wang}]{datasurvey}
Albalak, A.; Elazar, Y.; Xie, S.~M.; Longpre, S.; Lambert, N.; Wang, X.; Muennighoff, N.; Hou, B.; Pan, L.; Jeong, H.; Raffel, C.; Chang, S.; Hashimoto, T.; and Wang, W.~Y. 2024.
\newblock A Survey on Data Selection for Language Models.
\newblock \emph{Transactions on Machine Learning Research}.
\newblock Survey Certification.

\bibitem[{Bisk et~al.(2020)Bisk, Zellers, Gao, Choi et~al.}]{bisk2020piqa}
Bisk, Y.; Zellers, R.; Gao, J.; Choi, Y.; et~al. 2020.
\newblock Piqa: Reasoning about physical commonsense in natural language.
\newblock In \emph{Proceedings of the AAAI conference on artificial intelligence}, volume~34, 7432--7439.

\bibitem[{Blei, Ng, and Jordan(2003)}]{blei2003latent}
Blei, D.~M.; Ng, A.~Y.; and Jordan, M.~I. 2003.
\newblock Latent dirichlet allocation.
\newblock \emph{Journal of machine Learning research}, 3(Jan): 993--1022.

\bibitem[{Clark et~al.(2018)Clark, Cowhey, Etzioni, Khot, Sabharwal, Schoenick, and Tafjord}]{arc}
Clark, P.; Cowhey, I.; Etzioni, O.; Khot, T.; Sabharwal, A.; Schoenick, C.; and Tafjord, O. 2018.
\newblock Think you have solved question answering? try arc, the ai2 reasoning challenge.
\newblock \emph{arXiv preprint arXiv:1803.05457}.

\bibitem[{Devlin et~al.(2019)Devlin, Chang, Lee, and Toutanova}]{bert}
Devlin, J.; Chang, M.-W.; Lee, K.; and Toutanova, K. 2019.
\newblock {BERT}: Pre-training of Deep Bidirectional Transformers for Language Understanding.
\newblock In \emph{Proceedings of the 2019 Conference of the North {A}merican Chapter of the Association for Computational Linguistics: Human Language Technologies, Volume 1 (Long and Short Papers)}, 4171--4186. Minneapolis, Minnesota: Association for Computational Linguistics.

\bibitem[{Dubey et~al.(2024)Dubey, Jauhri, Pandey, Kadian, Al-Dahle, Letman, Mathur, Schelten, Yang, Fan et~al.}]{dubey2024llama}
Dubey, A.; Jauhri, A.; Pandey, A.; Kadian, A.; Al-Dahle, A.; Letman, A.; Mathur, A.; Schelten, A.; Yang, A.; Fan, A.; et~al. 2024.
\newblock The llama 3 herd of models.
\newblock \emph{arXiv preprint arXiv:2407.21783}.

\bibitem[{Fan, Pagliardini, and Jaggi(2024)}]{doge}
Fan, S.; Pagliardini, M.; and Jaggi, M. 2024.
\newblock {DOGE}: Domain Reweighting with Generalization Estimation.
\newblock In \emph{Forty-first International Conference on Machine Learning}.

\bibitem[{Gao et~al.(2023)Gao, Tow, Abbasi, Biderman, Black, DiPofi, Foster, Golding, Hsu, Le~Noac'h, Li, McDonell, Muennighoff, Ociepa, Phang, Reynolds, Schoelkopf, Skowron, Sutawika, Tang, Thite, Wang, Wang, and Zou}]{eval-harness}
Gao, L.; Tow, J.; Abbasi, B.; Biderman, S.; Black, S.; DiPofi, A.; Foster, C.; Golding, L.; Hsu, J.; Le~Noac'h, A.; Li, H.; McDonell, K.; Muennighoff, N.; Ociepa, C.; Phang, J.; Reynolds, L.; Schoelkopf, H.; Skowron, A.; Sutawika, L.; Tang, E.; Thite, A.; Wang, B.; Wang, K.; and Zou, A. 2023.
\newblock A framework for few-shot language model evaluation.

\bibitem[{Ge et~al.(2025)Ge, Huang, Cooper, Trost, Chu, GNVV, Cai, Park, Roberts, and Sala}]{ge2025r}
Ge, A.; Huang, T.-H.; Cooper, J.; Trost, A.; Chu, Z.; GNVV, S. S. S.~N.; Cai, Z.; Park, K.; Roberts, N.; and Sala, F. 2025.
\newblock R\&B: Domain Regrouping and Data Mixture Balancing for Efficient Foundation Model Training.
\newblock \emph{arXiv preprint arXiv:2505.00358}.

\bibitem[{Grootendorst(2020)}]{grootendorst2020bertopic}
Grootendorst, M. 2020.
\newblock BERTopic: Leveraging BERT and c-TF-IDF to create easily interpretable topics.
\newblock \emph{Zenodo, Version v0}, 9(10.5281).

\bibitem[{Johnson, Douze, and J{\'e}gou(2019)}]{johnson2019billion}
Johnson, J.; Douze, M.; and J{\'e}gou, H. 2019.
\newblock Billion-scale similarity search with {GPUs}.
\newblock \emph{IEEE Transactions on Big Data}, 7(3): 535--547.

\bibitem[{Lai et~al.(2017)Lai, Xie, Liu, Yang, and Hovy}]{race}
Lai, G.; Xie, Q.; Liu, H.; Yang, Y.; and Hovy, E. 2017.
\newblock {RACE}: Large-scale {R}e{A}ding Comprehension Dataset From Examinations.
\newblock In \emph{Proceedings of the 2017 Conference on Empirical Methods in Natural Language Processing}, 785--794. Copenhagen, Denmark: Association for Computational Linguistics.

\bibitem[{Li et~al.(2024)Li, Fang, Smyrnis, Ivgi, Jordan, Gadre, Bansal, Guha, Keh, Arora, Garg, Xin, Muennighoff, Heckel, Mercat, Chen, Gururangan, Wortsman, Albalak, Bitton, Nezhurina, Abbas, Hsieh, Ghosh, Gardner, Kilian, Zhang, Shao, Pratt, Sanyal, Ilharco, Daras, Marathe, Gokaslan, Zhang, Chandu, Nguyen, Vasiljevic, Kakade, Song, Sanghavi, Faghri, Oh, Zettlemoyer, Lo, {El-Nouby}, Pouransari, Toshev, Wang, Groeneveld, Soldaini, Koh, Jitsev, Kollar, Dimakis, Carmon, Dave, Schmidt, and Shankar}]{DCLM2024}
Li, J.; Fang, A.; Smyrnis, G.; Ivgi, M.; Jordan, M.; Gadre, S.; Bansal, H.; Guha, E.; Keh, S.; Arora, K.; Garg, S.; Xin, R.; Muennighoff, N.; Heckel, R.; Mercat, J.; Chen, M.; Gururangan, S.; Wortsman, M.; Albalak, A.; Bitton, Y.; Nezhurina, M.; Abbas, A.; Hsieh, C.-Y.; Ghosh, D.; Gardner, J.; Kilian, M.; Zhang, H.; Shao, R.; Pratt, S.; Sanyal, S.; Ilharco, G.; Daras, G.; Marathe, K.; Gokaslan, A.; Zhang, J.; Chandu, K.; Nguyen, T.; Vasiljevic, I.; Kakade, S.; Song, S.; Sanghavi, S.; Faghri, F.; Oh, S.; Zettlemoyer, L.; Lo, K.; {El-Nouby}, A.; Pouransari, H.; Toshev, A.; Wang, S.; Groeneveld, D.; Soldaini, L.; Koh, P.~W.; Jitsev, J.; Kollar, T.; Dimakis, A.~G.; Carmon, Y.; Dave, A.; Schmidt, L.; and Shankar, V. 2024.
\newblock {{DataComp-LM}}: {{In}} Search of the next Generation of Training Sets for Language Models.
\newblock arXiv:2406.11794.

\bibitem[{Liu et~al.(2024)Liu, Zheng, Muennighoff, Zeng, Dou, Pang, Jiang, and Lin}]{regmix}
Liu, Q.; Zheng, X.; Muennighoff, N.; Zeng, G.; Dou, L.; Pang, T.; Jiang, J.; and Lin, M. 2024.
\newblock RegMix: Data Mixture as Regression for Language Model Pre-training.
\newblock \emph{arXiv preprint arXiv:2407.01492}.

\bibitem[{Longpre et~al.(2024)Longpre, Yauney, Reif, Lee, Roberts, Zoph, Zhou, Wei, Robinson, Mimno, and Ippolito}]{pretrainersguide}
Longpre, S.; Yauney, G.; Reif, E.; Lee, K.; Roberts, A.; Zoph, B.; Zhou, D.; Wei, J.; Robinson, K.; Mimno, D.; and Ippolito, D. 2024.
\newblock A Pretrainer{'}s Guide to Training Data: Measuring the Effects of Data Age, Domain Coverage, Quality, {\&} Toxicity.
\newblock In Duh, K.; Gomez, H.; and Bethard, S., eds., \emph{Proceedings of the 2024 Conference of the North American Chapter of the Association for Computational Linguistics: Human Language Technologies (Volume 1: Long Papers)}, 3245--3276. Mexico City, Mexico: Association for Computational Linguistics.

\bibitem[{Meng et~al.(2015)Meng, Bradley, Yavuz, Sparks, Venkataraman, Liu, Freeman, Tsai, Amde, Owen, Xin, Xin, Franklin, Zadeh, Zaharia, and Talwalkar}]{meng2015mllibmachinelearningapache}
Meng, X.; Bradley, J.; Yavuz, B.; Sparks, E.; Venkataraman, S.; Liu, D.; Freeman, J.; Tsai, D.; Amde, M.; Owen, S.; Xin, D.; Xin, R.; Franklin, M.~J.; Zadeh, R.; Zaharia, M.; and Talwalkar, A. 2015.
\newblock MLlib: Machine Learning in Apache Spark.
\newblock arXiv:1505.06807.

\bibitem[{Mihaylov et~al.(2018)Mihaylov, Clark, Khot, and Sabharwal}]{openbookqa}
Mihaylov, T.; Clark, P.; Khot, T.; and Sabharwal, A. 2018.
\newblock Can a Suit of Armor Conduct Electricity? A New Dataset for Open Book Question Answering.
\newblock In \emph{Proceedings of the 2018 Conference on Empirical Methods in Natural Language Processing}, 2381--2391. Brussels, Belgium: Association for Computational Linguistics.

\bibitem[{Mu et~al.(2024{\natexlab{a}})Mu, Bai, Bontcheva, and Song}]{mu2024addressing}
Mu, Y.; Bai, P.; Bontcheva, K.; and Song, X. 2024{\natexlab{a}}.
\newblock Addressing Topic Granularity and Hallucination in Large Language Models for Topic Modelling.
\newblock \emph{arXiv preprint arXiv:2405.00611}.

\bibitem[{Mu et~al.(2024{\natexlab{b}})Mu, Dong, Bontcheva, and Song}]{mu2024large}
Mu, Y.; Dong, C.; Bontcheva, K.; and Song, X. 2024{\natexlab{b}}.
\newblock Large Language Models Offer an Alternative to the Traditional Approach of Topic Modelling.
\newblock \emph{arXiv preprint arXiv:2403.16248}.

\bibitem[{Parmar et~al.(2024)Parmar, Prabhumoye, Jennings, Liu, Jhunjhunwala, Wang, Patwary, Shoeybi, and Catanzaro}]{dataeverywhere}
Parmar, J.; Prabhumoye, S.; Jennings, J.; Liu, B.; Jhunjhunwala, A.; Wang, Z.; Patwary, M.; Shoeybi, M.; and Catanzaro, B. 2024.
\newblock Data, Data Everywhere: A Guide for Pretraining Dataset Construction.
\newblock In Al-Onaizan, Y.; Bansal, M.; and Chen, Y.-N., eds., \emph{Proceedings of the 2024 Conference on Empirical Methods in Natural Language Processing}, 10671--10695. Miami, Florida, USA: Association for Computational Linguistics.

\bibitem[{Penedo et~al.(2024)Penedo, Kydl{\'\i}{\v{c}}ek, Lozhkov, Mitchell, Raffel, Von~Werra, Wolf et~al.}]{fineweb}
Penedo, G.; Kydl{\'\i}{\v{c}}ek, H.; Lozhkov, A.; Mitchell, M.; Raffel, C.; Von~Werra, L.; Wolf, T.; et~al. 2024.
\newblock The FineWeb Datasets: Decanting the Web for the Finest Text Data at Scale.
\newblock \emph{arXiv preprint arXiv:2406.17557}.

\bibitem[{Rae et~al.(2021)Rae, Borgeaud, Cai, Millican, Hoffmann, Song, Aslanides, Henderson, Ring, Young et~al.}]{gopher}
Rae, J.~W.; Borgeaud, S.; Cai, T.; Millican, K.; Hoffmann, J.; Song, F.; Aslanides, J.; Henderson, S.; Ring, R.; Young, S.; et~al. 2021.
\newblock Scaling language models: Methods, analysis \& insights from training gopher.
\newblock \emph{arXiv preprint arXiv:2112.11446}.

\bibitem[{Rijcken et~al.(2023)Rijcken, Scheepers, Zervanou, Spruit, Mosteiro, and Kaymak}]{rijcken2023towards}
Rijcken, E.; Scheepers, F.; Zervanou, K.; Spruit, M.; Mosteiro, P.; and Kaymak, U. 2023.
\newblock Towards interpreting topic models with ChatGPT.
\newblock In \emph{The 20th World Congress of the International Fuzzy Systems Association}.

\bibitem[{Sakaguchi et~al.(2020)Sakaguchi, Le~Bras, Bhagavatula, and Choi}]{winogrande}
Sakaguchi, K.; Le~Bras, R.; Bhagavatula, C.; and Choi, Y. 2020.
\newblock WinoGrande: An Adversarial Winograd Schema Challenge at Scale.
\newblock In \emph{Proceedings of the AAAI Conference on Artificial Intelligence}, volume~34, 8732--8740.

\bibitem[{Sap et~al.(2019)Sap, Rashkin, Chen, Le~Bras, and Choi}]{siqa}
Sap, M.; Rashkin, H.; Chen, D.; Le~Bras, R.; and Choi, Y. 2019.
\newblock Social {IQ}a: Commonsense Reasoning about Social Interactions.
\newblock In \emph{Proceedings of the 2019 Conference on Empirical Methods in Natural Language Processing and the 9th International Joint Conference on Natural Language Processing (EMNLP-IJCNLP)}, 4463--4473. Hong Kong, China: Association for Computational Linguistics.

\bibitem[{Soboleva et~al.(2023)Soboleva, Al-Khateeb, Myers, Steeves, Hestness, and Dey}]{slimpajama}
Soboleva, D.; Al-Khateeb, F.; Myers, R.; Steeves, J.~R.; Hestness, J.; and Dey, N. 2023.
\newblock {SlimPajama: A 627B token cleaned and deduplicated version of RedPajama}.

\bibitem[{Su et~al.(2024)Su, Ahmed, Lu, Pan, Bo, and Liu}]{su2024roformer}
Su, J.; Ahmed, M.; Lu, Y.; Pan, S.; Bo, W.; and Liu, Y. 2024.
\newblock Roformer: Enhanced transformer with rotary position embedding.
\newblock \emph{Neurocomputing}, 568: 127063.

\bibitem[{Talmor et~al.(2019)Talmor, Herzig, Lourie, and Berant}]{commonsenseqa}
Talmor, A.; Herzig, J.; Lourie, N.; and Berant, J. 2019.
\newblock {C}ommonsense{QA}: A Question Answering Challenge Targeting Commonsense Knowledge.
\newblock In \emph{Proceedings of the 2019 Conference of the North {A}merican Chapter of the Association for Computational Linguistics: Human Language Technologies, Volume 1 (Long and Short Papers)}, 4149--4158. Minneapolis, Minnesota: Association for Computational Linguistics.

\bibitem[{Touvron et~al.(2023)Touvron, Lavril, Izacard, Martinet, Lachaux, Lacroix, Rozi{\`e}re, Goyal, Hambro, Azhar et~al.}]{llama}
Touvron, H.; Lavril, T.; Izacard, G.; Martinet, X.; Lachaux, M.-A.; Lacroix, T.; Rozi{\`e}re, B.; Goyal, N.; Hambro, E.; Azhar, F.; et~al. 2023.
\newblock Llama: Open and efficient foundation language models.
\newblock \emph{arXiv preprint arXiv:2302.13971}.

\bibitem[{Welbl, Liu, and Gardner(2017)}]{sciq}
Welbl, J.; Liu, N.~F.; and Gardner, M. 2017.
\newblock Crowdsourcing Multiple Choice Science Questions.
\newblock In \emph{Proceedings of the 3rd Workshop on Noisy User-generated Text}, 94--106. Copenhagen, Denmark: Association for Computational Linguistics.

\bibitem[{Wettig et~al.(2024)Wettig, Gupta, Malik, and Chen}]{QuRating}
Wettig, A.; Gupta, A.; Malik, S.; and Chen, D. 2024.
\newblock QuRating: Selecting High-Quality Data for Training Language Models.
\newblock In \emph{Forty-first International Conference on Machine Learning}.

\bibitem[{Wettig et~al.(2025)Wettig, Lo, Min, Hajishirzi, Chen, and Soldaini}]{WebOrganizer}
Wettig, A.; Lo, K.; Min, S.; Hajishirzi, H.; Chen, D.; and Soldaini, L. 2025.
\newblock Organize the {{Web}}: {{Constructing Domains Enhances Pre-Training Data Curation}}.
\newblock \emph{arXiv:2502.10341}.

\bibitem[{Xiao et~al.(2023)Xiao, Liu, Zhang, and Muennighoff}]{bge_embedding}
Xiao, S.; Liu, Z.; Zhang, P.; and Muennighoff, N. 2023.
\newblock C-Pack: Packaged Resources To Advance General Chinese Embedding.
\newblock arXiv:2309.07597.

\bibitem[{Xie et~al.(2023)Xie, Pham, Dong, Du, Liu, Lu, Liang, Le, Ma, and Yu}]{doremi}
Xie, S.~M.; Pham, H.; Dong, X.; Du, N.; Liu, H.; Lu, Y.; Liang, P.~S.; Le, Q.~V.; Ma, T.; and Yu, A.~W. 2023.
\newblock DoReMi: Optimizing Data Mixtures Speeds Up Language Model Pretraining.
\newblock In \emph{Advances in Neural Information Processing Systems}, volume~36, 69798--69818. Curran Associates, Inc.

\bibitem[{Zhuang et~al.(2025)Zhuang, Peng, Ma, Wang, Bai, Wei, Qiu, Zhang, Qian, and He}]{meta-rater}
Zhuang, X.; Peng, J.; Ma, R.; Wang, Y.; Bai, T.; Wei, X.; Qiu, J.; Zhang, C.; Qian, Y.; and He, C. 2025.
\newblock Meta-rater: A Multi-dimensional Data Selection Method for Pre-training Language Models.
\newblock \emph{arXiv preprint arXiv:2504.14194}.

\end{thebibliography}

\clearpage
\newpage

\appendix

\section{Case Study of Topic Extraction}
\label{app:topic_extraction_case_study}

\begin{table}[!tb]
    \small
    \caption{Examples across different topic extraction stages.}
    \centering
    \begin{tabular}{@{}lll@{}}
    \toprule
     \textbf{Final Topic} & \textbf{Topic} \\
     (12 items) &  (300 items) \\ \midrule
    Law & \begin{tabular}[l]{@{}l@{}} Traffic Safety and Regulations \\ Legal Regulations and Governance\end{tabular} \\ \midrule
    Lifestyle & \begin{tabular}[l]{@{}l@{}} Asian Cuisine and Healthy Recipes \\ Fashion and Beauty Industry\end{tabular}\\ \midrule
    Entertainment & \begin{tabular}[l]{@{}l@{}} Anime and Manga Culture \\ Celebrity Culture and Controversies\end{tabular}\\ \midrule
    Finance & \begin{tabular}[l]{@{}l@{}} Accounting and Taxation \\ Job Market and Career Opportunities \end{tabular}\\ \midrule
    Community & \begin{tabular}[l]{@{}l@{}}Religion and Cultural Practices \\ Community News and Local Affairs\end{tabular}\\ \midrule
    Science & \begin{tabular}[l]{@{}l@{}}Mathematical Analysis and Calculus \\ Geography and Demographics\end{tabular}\\ \midrule
    Health & \begin{tabular}[l]{@{}l@{}}Food Security and Nutrition\\ Public Health and Infectious Diseases\end{tabular}  \\ \midrule
    Education & \begin{tabular}[l]{@{}l@{}}Education and Career Development \\ Literature and Writing\end{tabular}  \\ \midrule
    Technology & \begin{tabular}[l]{@{}l@{}}Cloud Services and Cybersecurity\\ iOS Development and Programming\end{tabular}\\ \midrule
    Politics & \begin{tabular}[l]{@{}l@{}}African Political and Social Issues\\ International Relations and Geopolitics\end{tabular}\\ \midrule
    Relationships & \begin{tabular}[l]{@{}l@{}}Online Dating and Relationships\\ Relationships and Personal Growth\end{tabular}\\ \midrule
    Others & Miscellaneous\\      \bottomrule
    \end{tabular}
    \label{tab:dataweave_process}
\end{table}

\begin{figure*}[!t]  
    \centering
    \includegraphics[width=0.95\textwidth]{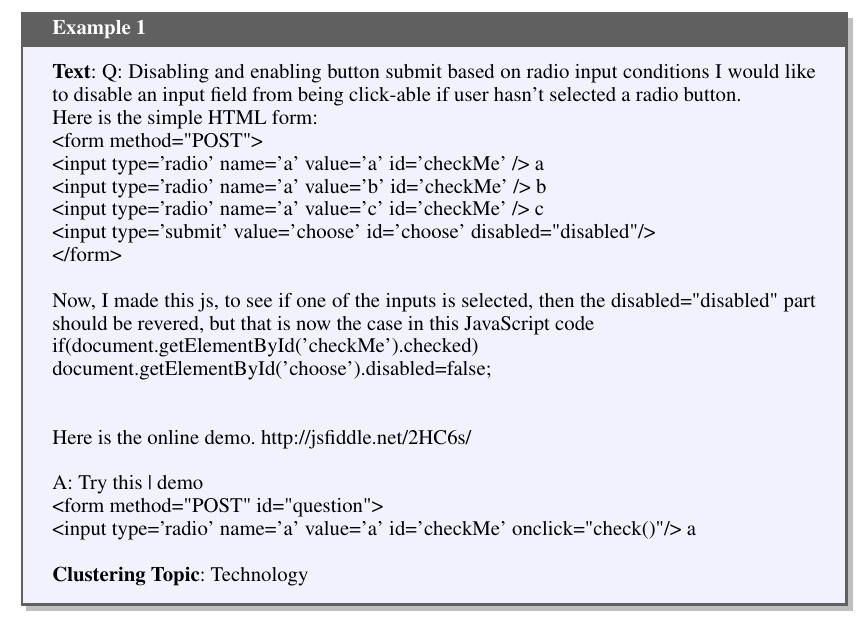} 
    \caption{Topic extraction example 1.}
    \label{fig:ex1}
\end{figure*}

\begin{figure*}[!t]  
    \centering
    \includegraphics[width=0.95\textwidth]{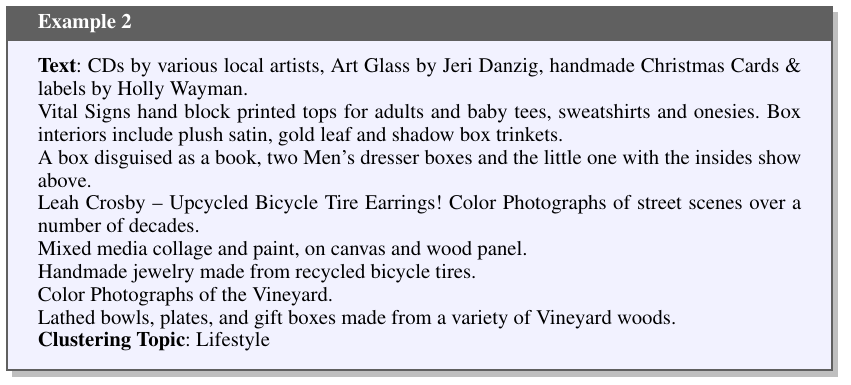} 
    \caption{Topic extraction example 2.}
    \label{fig:ex2}
\end{figure*}

\begin{figure*}[!t]  
    \centering
    \includegraphics[width=0.95\textwidth]{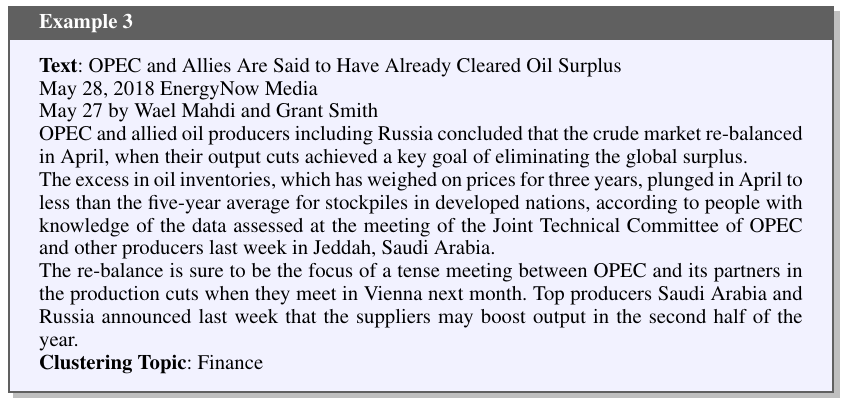} 
    \caption{Topic extraction example 3.}
    \label{fig:ex3}
\end{figure*}

\begin{figure*}[!t]  
    \centering
    \includegraphics[width=0.95\textwidth]{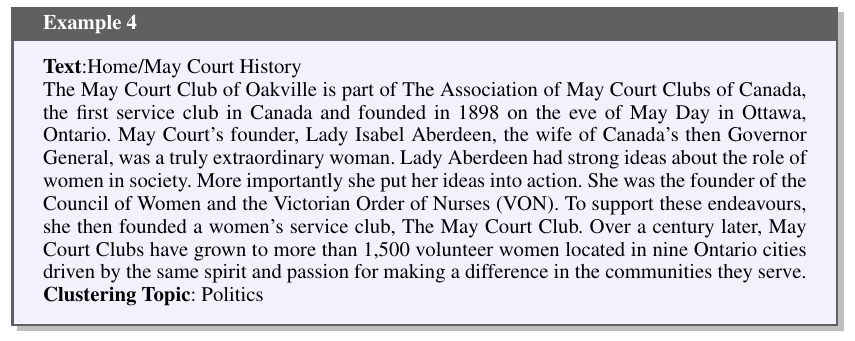} 
    \caption{Topic extraction example 4.}
    \label{fig:ex4}
\end{figure*}

Table \ref{tab:dataweave_process} presents several examples in the topic extraction process, illustrating the progression from 10,000 summaries to 300 identified topics, ultimately distilled into 12 final topics. To better showcase the results of extracted topics, we have selected some examples for demonstration, as shown in Figures \ref{fig:ex1}, \ref{fig:ex2}, \ref{fig:ex3}, and \ref{fig:ex4}.

\paragraph{LLMs can extract high-quality topics from summaries.}
Unlike individual words, summaries encapsulate information from multiple documents, providing a rich semantic foundation for topic extraction. 
This complexity allows LLMs to identify and extract high-quality, human-readable topics from these summaries effectively. 
The ability of LLMs to synthesize and distill nuanced themes underscores their potential in various NLP tasks, particularly in generating coherent and relevant topics that reflect the underlying content.

\paragraph{Merging topics is vital.}
\label{app:merging_topics}
The analysis reveals a notable issue of non-parallel topic granularity among the initial 300 human-interpretable topics. 
For example, the topic \textit{Gaming and Entertainment Overview} serves as a specific subset within the broader category of \textit{Entertainment}, while \textit{Jewelry and Timepieces} and \textit{Fashion and Beauty Industry} exhibit partitial overlap in the concepts. 
This discrepancy highlights the need for a systematic merging process to ensure clarity and coherence in topic categorization. 
Fortunately, this granularity issue has been effectively resolved in the final set of 12 topics, demonstrating the importance of refining topic definitions and relationships to enhance interpretability and usability in downstream applications.

\section{Prompt Templates} 
\label{app:prompt}
\begin{figure*}[!t]  
    \centering
    \includegraphics[width=0.95\textwidth]{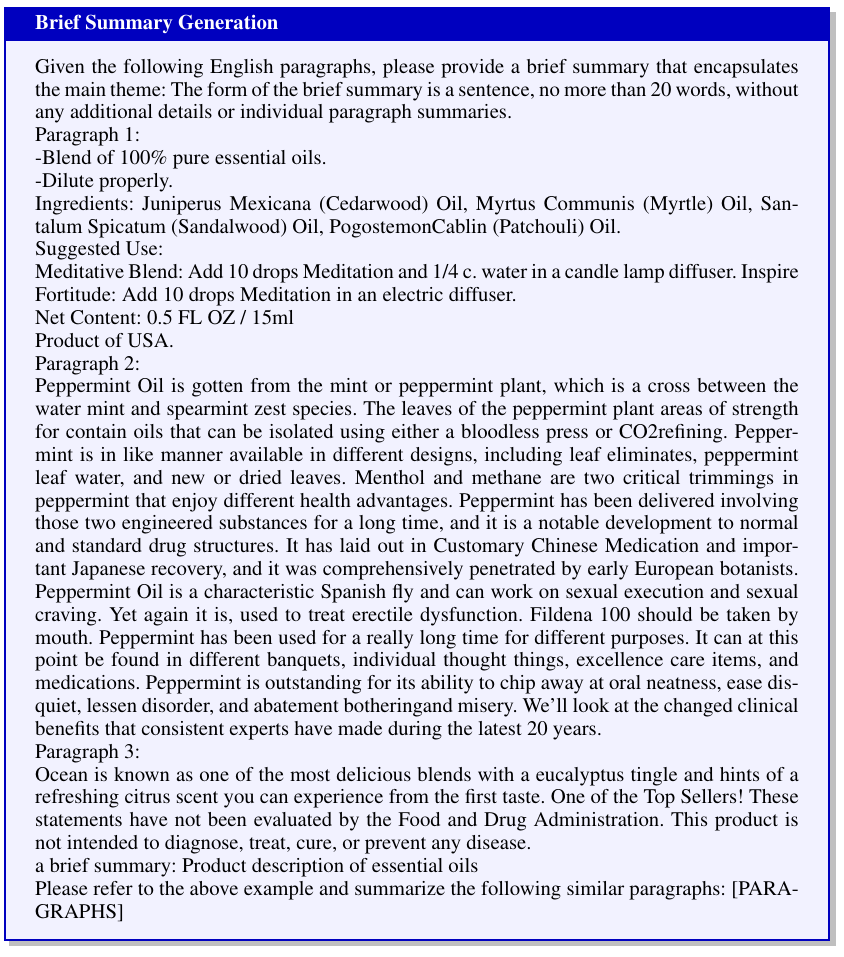} 
    \caption{The prompt of extracting brief summary for each partition.}
    \label{fig:prompt_brief_summary}
\end{figure*}

\begin{figure*}[!t]  
    \centering
    \includegraphics[width=0.95\textwidth]{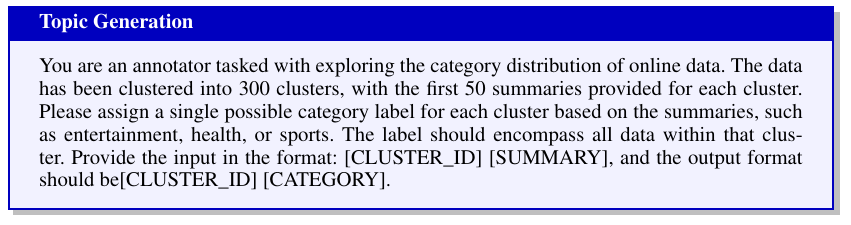} 
    \caption{The prompt of extracting summary to topic.}
    \label{fig:prompt_summary2prompt}
\end{figure*}

\begin{figure*}[!t]  
    \centering
    \includegraphics[width=0.95\textwidth]{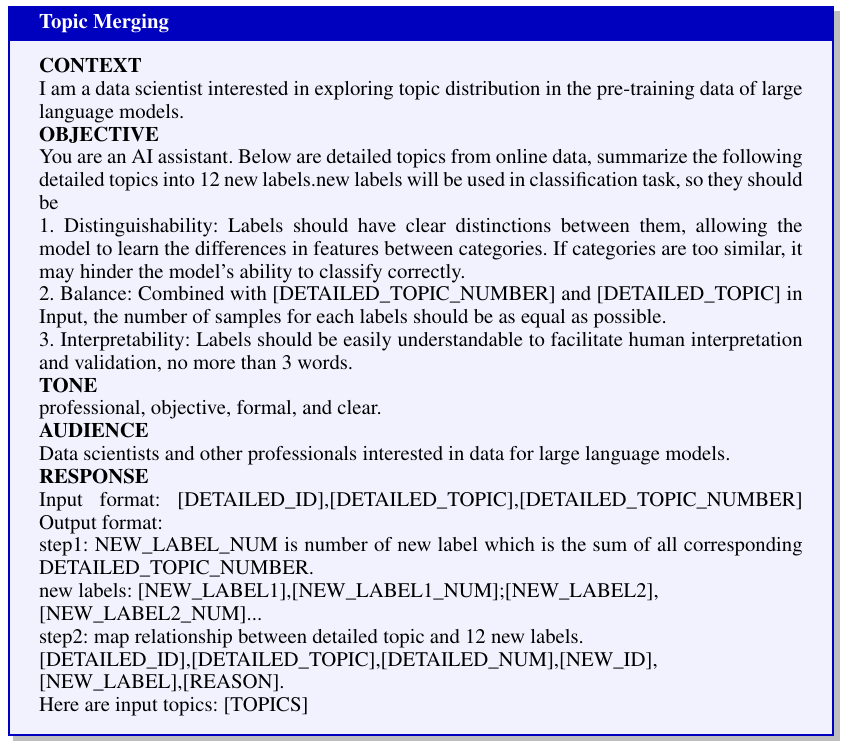} 
    \caption{The prompt of merging topics to final topics.}
    \label{fig:prompt_topic2topic}
\end{figure*}

We present three prompts utilized in our method, including generating a brief summary, deriving topics from summaries, and producing final topics. 
These prompts are illustrated in Figures \ref{fig:prompt_brief_summary}, \ref{fig:prompt_summary2prompt}, and \ref{fig:prompt_topic2topic}. 
We employ gpt-4o  to obtain the corresponding results.

\section{Training Details}
\label{app:training}

\subsection{Topic Classifier Training}
\label{app:topic_classifier}

We fine-tuned BERT for topic classification. 
The training dataset for topic classifier is derived from a subset of SlimPajama, comprising a total of 100,000 samples, which were divided into training, development, and test sets in a ratio of 8:1:1. 
The training process required approximately 8 NVIDIA A800 GPU hours for 10 epochs. 
Upon completion of the training, the topic classifier attained an accuracy score of 84\% on the test set.

\subsection{Pre-training}
\label{app:pretrain_detail}
The architecture of the pre-trained models are detailed in Table \ref{tab:architecture}. 
Each model was trained on 32x NVIDIA A800 GPUs, utilizing a global batch size of $4 \times 2^{20}$ tokens and completing 7,500 and 17,500 training steps within approximately 14 hours and 65 hours, respectively. 
The learning rate was set to $5 \times 10^{-5}$, and the Adam optimizer was used with the following hyperparameters: $\beta_1 = 0.9$, $\beta_2 = 0.95$, and $\epsilon = 10^{-8}$.

\begin{table}[htbp]
\centering
\begin{tabular}{@{}lcc@{}}
\toprule
\textbf{Hyperparameter} & \textbf{1.3B} & \textbf{3.3B} \\ \midrule
Hidden Dimension Size                       & 2,048         & 2,560         \\
Number of Layers                            & 24            & 40            \\
Number of Attention Heads                   & 16            & 20            \\
Number of KV Heads                          & 16            & 20            \\
Number of Total Parameters                  & 1,345,423,360 & 3,335,989,760 \\
Consumed Tokens (B)                         & 30            & 60            \\
Pre-training Time (h)                       & 14.0          & 60.0          \\ \bottomrule
\end{tabular}
\caption{The architecture of pre-trained decoder-only model.}
\label{tab:architecture}
\end{table}

\section{Data Weights for All Data Mixing Settings}
\label{app:all_mixing_weights}
The detailed source/topic weights in different settings are provided in Table \ref{tab:weights}. 

\paragraph{RegMix implementation.} Our RegMix approach followed the official implementation, where we generated 512 random domain mixtures for each defined domain. These mixtures were used to train 25M-parameter proxy models, each structured with a hidden dimension of 256, 4 layers, and 4 attention heads. Each proxy model was trained for 1000 steps on 8 NVIDIA A800 GPUs with a global batch size of $1 \times 2^{20}$ tokens.

\paragraph{DoReMi implementation.} Our DoReMi implementation is based on the official source code, using a 100M model trained on 7.5B tokens as the reference. During the iterative proxy training stage, domain weights were updated 30 times in total. The resulting weights were then applied to train a final 1.3B parameter model.

\begin{table*}[htbp]
\centering
\begin{tabular}{@{}lccccccc@{}}
\toprule
Topic/Source  & SlimPajama & RegMix & Temperature & PerfRe & DoReMi  \\ \midrule
Technology    & 17.55      & 14.91  & 10.35       & 13.50  &  17.37  \\
Science       & 5.73       & 5.54   & 7.70        & 12.10  &  5.77   \\
Politics      & 8.23       & 4.06   & 8.20        & 6.33   &  8.21   \\
Health        & 7.04       & 5.31   & 7.96        & 13.10  &  6.97   \\
Lifestyle     & 5.49       & 12.01  & 7.66        & 4.22   &  5.56   \\
Law           & 6.08       & 4.12   & 7.77        & 4.68   &  6.04   \\
Entertainment & 23.91      & 29.14  & 12.13       & 18.39  &  23.71  \\
Education     & 13.40      & 9.14   & 9.33        & 10.31  &  13.23  \\
Relationships & 1.14       & 6.16   & 6.87        & 8.57   &  1.34   \\
Finance       & 4.01       & 2.63   & 7.38        & 3.09   &  4.08   \\
Community     & 2.29       & 1.89   & 7.07        & 1.76   &  2.44   \\
Others        & 5.13       & 5.10   & 7.59        & 3.95   &  5.28   \\ \midrule
arXiv         & 4.60       & 4.04   & 10.40       & 3.50   & 4.71  \\
Book          & 4.20       & 6.35   & 10.33       & 3.20   & 4.48  \\
C4            & 26.70      & 62.83  & 17.24       & 32.10  & 26.46 \\
CommonCrawl   & 52.20      & 12.97  & 32.62       & 51.70  & 51.01 \\
Github        & 5.20       & 1.03   & 10.07       & 4.00   & 5.40  \\
StackExchange & 3.30       & 8.88   & 9.61        & 2.50   & 3.66  \\
Wikipedia     & 3.80       & 3.90   & 9.73        & 2.90   & 4.28  \\ \bottomrule
\end{tabular}
\caption{Exact Topic/Source weights (\%) on SlimPajama obtained in data mixing methods.}
\label{tab:weights}
\end{table*}

\section{Evaluation Details}
\label{app:evaluation}

We evaluated LLM performance under few-shot ICL settings using the lm-evaluation-harness framework for comprehensive comparison. 
Details for each downstream task are shown in Table \ref{tab:icl}.

\begin{table*}[htbp]
\centering
\begin{tabular}{@{}ccc@{}}
\toprule
\textbf{Task}                                   & \textbf{Dataset}       & \textbf{Number of Examples} \\ \midrule
\multirow{3}{*}{General Knowledge}     & ARC-E         & 15                 \\
                                       & ARC-C         & 15                 \\
                                       & SciQ          & 2                  \\ \midrule
\multirow{4}{*}{Commonsense Reasoning} & SIQA          & 10                 \\
                                       & PIQA          & 10                 \\
                                       & WinoGrande    & 15                 \\
                                       & CommonsenseQA & 10                 \\ \midrule
\multirow{2}{*}{Reading Comprehension} & RACE          & 2                  \\
                                       & OpenbookQA    & 10                 \\ \bottomrule
\end{tabular}
\caption{ICL evaluation details in our experiment.}
\label{tab:icl}
\end{table*}

\section{Full Experimental Results}
\label{app:results}

\subsection{Continual Pre-training Results}
\label{app:continual_pretrain}
The full result of continual pre-training is shown in Table \ref{tab:downstream}.

\begin{table*}[htbp]
    \centering
    \begin{tabular}{@{}lcccc@{}}
    \toprule
    \multicolumn{1}{c}{\textbf{Upsampled Topic}} &
      \begin{tabular}[c]{@{}c@{}}\textbf{General Knowledge}\\ (3 tasks)\end{tabular} &
      \begin{tabular}[c]{@{}c@{}}\textbf{Commonsense Reasoning}\\ (4 tasks)\end{tabular} &
      \begin{tabular}[c]{@{}c@{}}\textbf{Reading Comprehension}\\ (2 tasks)\end{tabular} &
      \begin{tabular}[c]{@{}c@{}}\textbf{Average}\\ (9 tasks)\end{tabular} \\ \midrule
    Random        & 57.91          & 45.24 & 32.56          & 46.64 \phantom{\posdiff{0.00}}                 \\ \midrule
    Technology    & 58.80          & 46.13          & 33.22          & 47.49 \posdiff{0.21}           \\
    Science       & 61.55          & 46.23          & 33.87          & 48.59 \posdiff{1.95} \\
    Politics      & 57.97          & 45.77          & 33.50          & 47.11 \posdiff{0.47}          \\
    Health        & 59.01          & 46.21          & 33.56          & 47.67 \posdiff{1.03}          \\
    Lifestyle     & 57.69          & 47.34          & 33.12          & 47.60 \posdiff{0.96}          \\
    Law           & 58.16          & 45.75          & 33.79          & 47.23 \posdiff{0.59}           \\
    Entertainment & 57.90          & 45.26          & 34.43          & 46.40 \negdiff{0.24}         \\
    Education     & 59.09          & 46.26          & 33.15          & 47.62 \posdiff{0.98}         \\
    Relationships & 58.96          & 46.88 &       32.43             & 47.70 \posdiff{1.06}        \\
    Finance       & 56.47          & 45.72          & 32.70          & 46.41 \negdiff{0.23}         \\
    Community     & 58.15          & 46.12          & 33.15          & 47.31 \posdiff{0.67}          \\
    Others        & 58.27          & 45.95          & 34.00          & 47.40 \posdiff{0.76}          \\ \midrule
    arXiv         & 57.32          & 44.91          & 31.34          & 46.03 \negdiff{0.61}         \\ 
    Book          & 57.33          & 45.95          & 30.75          & 46.36 \negdiff{0.28}         \\ 
    C4            & 58.62          & 46.41          & 30.94          & 47.04 \posdiff{0.40}          \\ 
    CommonCrawl   & 58.10           & 45.73          & 31.92          & 46.79 \posdiff{0.15}          \\ 
    Github        & 55.80           & 45.73          & 31.22          & 46.20 \negdiff{0.44}          \\ 
    StackExchange & 57.56          & 45.70           & 31.37          & 46.47 \negdiff{0.17}         \\ 
    Wikipedia     & 57.56          & 45.46          & 30.45          & 46.06 \negdiff{0.58}          \\ \bottomrule
    \end{tabular}
    \caption{Downstream tasks results of continual pre-training settings. \textit{Random} denotes no any control over topic distribution of the 30B additional tokens. The \posdiffinline{x.xx} and \negdiffinline{x.xx} values indicate positive and negative differences compared to the Random baseline, respectively.}
    \label{tab:downstream}
\end{table*}

\subsection{Full Results of Pre-training Models Using Different Data Mixing Methods}
\label{app:full_pretrain}
The full results of pre-training experiment under different data mixing methods are shown in Tables \ref{tab:fullcontinual_1}, \ref{tab:fullcontinual_2}, and \ref{tab:fullcontinual_3}.

\begin{table*}[htbp]
\centering
\begin{tabular}{@{}lcccc@{}}
\toprule
\textbf{Data Mixing Method} & \textbf{ARC-E} & \textbf{ARC-C} & \textbf{SciQ} & \textbf{Average} \\ \midrule
Random             & 52.44 & 26.21 & 84.90 & 54.52 \\ \midrule
PerfRe-Source      & 53.41 & 26.79 & 85.60 & 55.27 \\
PerfRe-Topic       & 55.72 & 27.47 & 85.90 & 56.36 \\ \midrule
Temperature-Source & 51.18 & 25.94 & 83.80 & 53.64 \\
Temperature-Topic  & 53.87 & 27.30 & 85.70 & 55.62 \\ \midrule
RegMix-Source      & 51.81 & 25.60 & 83.90 & 53.77 \\
RegMix-Topic       & 51.26 & 26.71 & 85.20 & 54.39 \\ \midrule
DoReMi-Source      & 53.45 & 26.54 & 83.10 & 54.36\\
DoReMi-Topic       & 53.79 & 26.96 & 84.20 & 54.98\\ \midrule
3.3B Random        & 63.84 & 28.32 & 91.50 & 61.22 \\
3.3B RegMix-Source & 62.71 & 31.05 & 90.10 & 61.29 \\
3.3B RegMix-Topic  & 62.42 & 31.72 & 91.20 & 61.78 \\ \bottomrule
\end{tabular}
\caption{Full downstream tasks results of pre-training using different data mixing methods in General Knowledge.}
\label{tab:fullcontinual_1}
\end{table*}

\begin{table*}[htbp]
\centering
\begin{tabular}{@{}lccccc@{}}
\toprule
\textbf{Data Mixing Method} & \textbf{SIQA} & \textbf{PIQA} & \textbf{WinoGrande} & \textbf{CommonsenseQA} & \multicolumn{1}{l}{\textbf{Average}} \\ \midrule
Random             & 39.36 & 67.46 & 51.70 & 19.16 & 44.42 \\ \midrule
PerfRe-Source      & 40.79 & 68.34 & 52.72 & 21.54 & 45.85 \\
PerfRe-Topic       & 40.07 & 69.53 & 52.96 & 22.36 & 46.23 \\ \midrule
Temperature-Source & 39.46 & 67.79 & 53.83 & 20.80 & 45.47 \\
Temperature-Topic  & 40.43 & 68.50 & 52.57 & 18.35 & 44.96 \\ \midrule
RegMix-Source      & 40.12 & 70.08 & 51.62 & 21.13 & 45.74 \\
RegMix-Topic       & 40.74 & 69.53 & 52.17 & 21.38 & 45.96 \\ \midrule
DoReMi-Source      & 41.50 & 68.50 & 51.70 & 19.66 & 45.34\\
DoReMi-Topic       & 41.25 & 68.55 & 53.51 & 18.67 & 45.50\\ \midrule
3.3B Random        & 42.63 & 71.65 & 57.93 & 17.12 & 47.33 \\
3.3B RegMix-Source & 41.61 & 74.70 & 55.96 & 18.10 & 47.59 \\
3.3B RegMix-Topic  & 43.19 & 73.07 & 56.76 & 21.76 & 48.70 \\ \bottomrule
\end{tabular}
\caption{Full downstream tasks results of pre-training using different data mixing methods in Commonsense Reasoning.}
\label{tab:fullcontinual_2}
\end{table*}

\begin{table*}[htbp]
\centering
\begin{tabular}{@{}lccc@{}}
\toprule
\textbf{Data Mixing Method} & \textbf{RACE} & \textbf{OpenBookQA} & \multicolumn{1}{l}{\textbf{Average}} \\ \midrule
Random             & 21.34 & 28.80 & 25.07 \\ \midrule
PerfRe-Source      & 21.72 & 30.80 & 26.26 \\
PerfRe-Topic       & 21.44 & 31.60 & 26.52 \\ \midrule
Temperature-Source & 22.11 & 29.40 & 25.76 \\
Temperature-Topic  & 24.11 & 31.20 & 27.66 \\ \midrule
RegMix-Source      & 20.96 & 29.80 & 25.38 \\
RegMix-Topic       & 23.16 & 29.40 & 26.28 \\ \midrule
DoReMi-Source      & 25.17 & 29.20 & 27.19 \\
DoReMi-Topic       & 26.03 & 32.00 & 29.02 \\ \midrule
3.3B Random        & 33.49 & 34.20 & 33.85 \\
3.3B RegMix-Source & 33.40 & 34.00 & 33.70 \\
3.3B RegMix-Topic  & 33.06 & 35.00 & 34.03 \\ \bottomrule
\end{tabular}
\caption{Full downstream tasks results of pre-training using different data mixing methods in Reading Comprehension.}
\label{tab:fullcontinual_3}
\end{table*}

\subsection{Full Results of Further Analysis}
\label{app:full_ablation}
The full results of ablation experiment are shown in Tables \ref{tab:fullmixing_1}, \ref{tab:fullmixing_2}, and \ref{tab:fullmixing_3}.

\begin{table*}[htbp]
\centering
\begin{tabular}{@{}lcccc@{}}
\toprule
\textbf{Data Mixing Method} & \textbf{ARC-E} & \textbf{ARC-C} & \textbf{SciQ} & \textbf{Average} \\ \midrule
Random                & 52.44 & 26.21 & 84.90 & 54.52 \\ \midrule
RegMix-Source         & 51.81 & 25.60 & 83.90 & 53.77 \\
RegMix-Topic          & 51.26 & 26.71 & 85.20 & 54.39 \\ \midrule
RegMix-Topic * Source & 56.40 & 26.88 & 86.40 & 56.56 \\ \midrule
FineWeb-Edu           & 54.92 & 27.47 & 85.80 & 56.06 \\
\quad + RegMix-Source & 57.83 & 28.50 & 87.40 & 57.91 \\
\quad + RegMix-Topic  & 57.38 & 28.92 & 88.00 & 58.10 \\ \midrule
RegMix-Topic-7        & 54.92 & 27.47 & 85.80 & 56.06 \\ \bottomrule
\end{tabular}
\caption{Full downstream tasks results of ablation experiment in General Knowledge.}
\label{tab:fullmixing_1}
\end{table*}

\begin{table*}[htbp]
\centering
\begin{tabular}{@{}lccccc@{}}
\toprule
\textbf{Data Mixing Method} & \textbf{SIQA} & \textbf{PIQA} & \textbf{WinoGrande} & \multicolumn{1}{l}{\textbf{CommonsenseQA}} & \textbf{Average} \\ \midrule
Random                & 39.36 & 67.46 & 51.70 & 19.16 & 44.42 \\ \midrule
RegMix-Source         & 40.12 & 70.08 & 51.62 & 21.13 & 45.74 \\
RegMix-Topic          & 40.74 & 69.53 & 52.17 & 21.38 & 45.96 \\ \midrule
RegMix-Topic * Source & 39.92 & 69.75 & 51.38 & 19.16 & 45.05 \\ \midrule
FineWeb-Edu           & 40.17 & 69.48 & 52.25 & 20.23 & 45.53 \\
\quad + RegMix-Source & 40.28 & 70.24 & 52.49 & 21.05 & 46.02 \\
\quad + RegMix-Topic  & 40.74 & 70.26 & 52.89 & 20.16 & 46.01 \\
RegMix-Topic-7        & 40.17 & 69.48 & 52.25 & 20.23 & 45.53 \\ \bottomrule
\end{tabular}
\caption{Full downstream tasks results of ablation experiment in Commonsense Reasoning.}
\label{tab:fullmixing_2}
\end{table*}

\begin{table*}[htbp]
\centering
\begin{tabular}{@{}lccc@{}}
\toprule
\textbf{Data Mixing Method} & \textbf{RACE} & \textbf{OpenBookQA} & \textbf{Average} \\ \midrule
Random                      & 21.34         & 28.80               & 25.07            \\ \midrule
RegMix-Source               & 20.96         & 29.80               & 25.38            \\
RegMix-Topic                & 23.16         & 29.40               & 26.28            \\ \midrule
RegMix-Topic * Source       & 22.30         & 29.00               & 25.65            \\ \midrule
FineWeb-Edu                 & 21.63         & 28.80               & 25.22            \\
\quad + RegMix-Source       & 21.34         & 32.20               & 26.77            \\
\quad + RegMix-Topic        & 23.76         & 32.00               & 27.88            \\ \midrule
RegMix-Topic-7              & 21.63         & 28.80               & 25.22            \\ \bottomrule
\end{tabular}
\caption{Full downstream tasks results of ablation experiment in Reading Comprehension.}
\label{tab:fullmixing_3}
\end{table*}

\end{document}